\documentclass[11pt, a4paper, onecolumn, copyright, goog]{google}

\usepackage[authoryear, sort&compress, round]{natbib}
\usepackage{booktabs}
\usepackage{adjustbox}
\usepackage{tcolorbox}
\usepackage{tcolorbox}
\tcbuselibrary{breakable}
\usepackage{caption}
\usepackage{multirow}
\usepackage{wrapfig}

\uselogo{} 

\title{Training User Simulators with Latent Intrinsic Feedback}

\correspondingauthor{tareknaous@gatech.edu}

\author[*1,2]{Tarek Naous}
\author[1]{Craig Boutilier}
\author[2]{Wei Xu}
\author[1]{Ofer Meshi}

\affil[1]{\thepa{}{}}
\affil[2]{Georgia Institute of Technology}

\newcommand{\icon}[1]{\raisebox{-0.3ex}{\includegraphics[height=1em]{#1}}}

\usepackage{amssymb}
\usepackage{pifont}

\newcommand{\xmark}{\ding{55}} 

\let\cite\citep 

\begin{abstract}
User simulators are gaining attention as environments to train LLM assistants that can better cope with the nuances of multi-turn conversations with real users. Prior work on user simulation has emphasized generating realistic user utterances, leaving unexplored their potential as a source of direct feedback on the assistant's responses. Such evaluation signals may not be explicitly communicated in language, but can instead reside in the latent reasoning of users. In this work, we introduce \emph{user simulators with intrinsic feedback}, namely, models equipped with \emph{latent evaluation states}: internal thoughts, reactions, and scalar ratings that evaluate the assistant from a first-person user perspective at each conversation turn. We train LLM-based simulators using interaction logs between real users and assistants and demonstrate that intrinsic feedback improves simulation realism and critiquing behaviors versus prompted simulator baselines, while also providing turn-level evaluations that existing fine-tuned simulators lack. Further, evaluations generated by our trained simulators correlate better with human judgments than those of prompted simulators of the same size. Lastly, our experiments reveal that simulators with intrinsic feedback provide effective reward signals for multi-turn RL training, matching setups that rely on external judge LLMs for rewards while cutting RL compute by more than half.
\end{abstract}

\begin{document}

\maketitle

\section{Introduction}

AI agents are transforming the way people interact with digital platforms. Instead of doing tasks directly, users now spend ever-increasing time engaging in conversations with LLMs to get work done, unlocking new levels of productivity. According to Pew \cite{pew2026ai}, the use of AI chatbots by U.S.\ adults has been on the rise, with about half reporting daily usage as of June 2026, and 38\% using AI to complete tasks at work. Despite this wide adoption, recent work has highlighted shortcomings in how LLM assistants handle multi-turn conversations, such as making incorrect assumptions about unspecified details in earlier turns \cite{laban2026llms} or losing track of the user's goals \cite{tack2026llms}. In fact, analyses of user-assistant interactions by \citet{baumann2026swe} show that users push back against AI assistants' outputs 44\% of the time through corrections and interruptions. To cope with such challenges, training user simulators that replicate the nuanced behavior of human users is gaining increased attention \cite{naous2026flipping,zhou2026odyssim}.

Importantly, a key use of user simulators is to provide a realistic environment for multi-turn RL training of assistant LLMs \cite{wan2026enhancing,abdulhai2025evaluating}. Such RL training requires inferring the  value created for the simulated user
(e.g., degree of satisfaction)  to provide a reward signal for the assistant. Yet, current simulators are trained primarily as utterance generators, requiring external judge models to estimate this signal from interactions \cite{suh2026quantifying}. This is limiting, first, because external judges may poorly correlate with the subjective first-person user assessment \cite{meshi2026convapparel,dou2025simulatorarena}; and second, additional LLM-as-judge calls during training increase computational overhead significantly. A natural question thus emerges: \textit{can we build user simulators that provide direct evaluations of how well an assistant is serving them?}

To address this question, we propose \textbf{user simulators with intrinsic feedback}, which not only model \emph{what users say} but also \emph{how they evaluate} the assistant. A user's evaluation of an assistant's response, however, is not always explicitly expressed in their next utterance. For example, a user may recognize that a response is incomplete, but continue the conversation by reformulating or refining their request instead of pointing out the assistant's failure \cite{yuan2025query}. The assessment driving this behavior may therefore reside in the user's \textit{latent reasoning} rather than in their observable language \cite{wu2026humanlm,tennenholtz2026controllable}. Inspired by this phenomenon, we train user simulators that generate \textbf{latent evaluation states}, which augment the user simulator's latent reasoning with explicit feedback signals: the user's internal thoughts, reactions, feedback summaries, and scalar ratings that evaluate the assistant's response at each conversation turn. This better models the user's experience while providing direct, simultaneous evaluation of the assistant's quality by the user simulator itself.

We begin by training user simulators equipped with latent evaluation states based on Gemma3 models \cite{team2025gemma} (4B, 12B, and 27B parameters) using interaction logs between real users and LLMs (Sec.~\ref{sec:latent-eval-state-sims}). Our results show that user simulators trained with latent feedback achieve better simulation realism than standard baselines that either prompt instruction-tuned LLMs to role-play users or fine-tuned simulators without latent states (Sec.~\ref{sec:evaluations-intrinsic}). Further, we show that our simulators better capture how users critique assistant responses compared to prompted simulators, with their ratings correlating more strongly with human judgments on conversations with real user ratings.

Lastly, we demonstrate the value of leveraging the latent evaluations produced by our trained simulators for multi-turn RL training of assistants (Sec.~\ref{sec:rl-assistant-results}). We find that the enhanced realism of our trained simulators enables better assistant training than prompted simulator baselines. Further, the intrinsic evaluations produced by our simulators provide an effective reward signal, producing competitive assistant LLMs. Crucially, by eliminating the need for judge models to compute rewards, we reduce the computation required for an RL training run by \emph{more than half}, yielding a more compute-efficient training recipe.

\section{Background on Training User Simulators}
\label{sec:background}

User simulation has increasingly shifted toward training on real user interactions \cite{lu2026,kolluri2025}, rather than relying solely on prompting LLMs to act as users \cite{zhu2025llm,mehri2026goal}. We consider a setting where an LLM assistant $\mathcal{A}$ engages in a multi-turn conversation with a user simulator $\mathcal{U}$. Let $c_t = [u_1,a_1, ..., u_{t},a_{t}]$ denote the conversation history up to turn $t$. The goal of $\mathcal{U}$ is to model $P_{\mathcal{U}}(u_{t+1} | c_t)$, the probability of generating the next user turn $u_{t+1}$,
as an autoregressive mapping conditioned on $c_t$ (including the latest response from $\mathcal{A}$). $\mathcal{A}$ samples responses according to a learned policy $P_{\mathcal{A}}(a_{t+1} | c_t, u_{t+1})$. The probability of a conversation trajectory $\mathcal{T}$ with $N$ turns adheres to the causal decomposition:
$$P(\mathcal{T}) = \prod_{t=0}^{N-1} P_{\mathcal{U}}(u_{t+1} | c_t) P_{\mathcal{A}}(a_{t+1} | c_t, u_{t+1}).$$
To approximate $P_{\mathcal{U}}(u_{t+1} | c_t)$, the typical approach to training user simulators is \emph{supervised fine-tuning (SFT)} on interaction logs between real users and assistants. Given a dataset $\mathcal{D} = \{ \mathcal{T}_i\}_i^{|\mathcal{D}|}$ of conversational trajectories, a pretrained LLM parameterized by $\theta$ is fine-tuned to maximize the log-likelihood of these human utterances given  the conversational context: $\mathcal{L_{\mathrm{SFT}}} = - \mathbb{E}_{\mathcal{T\sim\mathcal{D}}} [\sum_{t=1}^N \log P_{\theta}(u_{t+1} | c_t)]$. By modeling this distribution autoregressively over tokens, the trained simulator learns the general behavior of real users. While such simulators can generate realistic user utterances, they are difficult to \emph{control}, i.e., to direct the conversation to reflect a particular user intent. This is a critical capability for user simulation, since it allows reliable simulation of specific tasks when evaluating or training assistants. An approach to address this was introduced by \citet{naous2026flipping} where the simulator is trained to model $P_{\mathcal{U}}(u_{t+1} | I, c_t)$ which conditions on the high-level user intent $I$ (i.e., what they are trying to achieve in this conversation), before the first user turn. Related approaches that train simulators conditioned on user profiles or personas include those of \citet{dou2025simulatorarena,wei2026mirroring, wang2025know}.

While intent conditioning can anchor simulators on specific tasks while maintaining realism, it ignores important, underlying
user dynamics at the turn-level.  This limitation has been tackled using \emph{latent states}: simulators trained with latent states model $P_{\mathcal{U}}(z_{t+1}, u_{t+1} | c_t) = P_{\mathcal{U}}(z_{t+1} | c_t) P_{\mathcal{U}}(u_{t+1} | z_{t+1}, c_t)$, where $z_t$ is an intermediate variable generated by the simulator before generating user turn $u_t$ \cite{wu2026humanlm}. Typically, $z_t$ is represented by a reasoning trace that is synthetically generated from empirical training data, aimed at aligning simulators with their pre-defined goals, values, or personas. Latent states have also been shown to make simulators more robust to assistants that differ from those observed in their training data \cite{tennenholtz2026controllable}.

\section{User Simulators with Latent Evaluation States}
\label{sec:latent-eval-state-sims}

\begin{wrapfigure}{r}{0.5\linewidth}
    \vspace{-1.0em} 
    \centering
    \includegraphics[width=\linewidth]{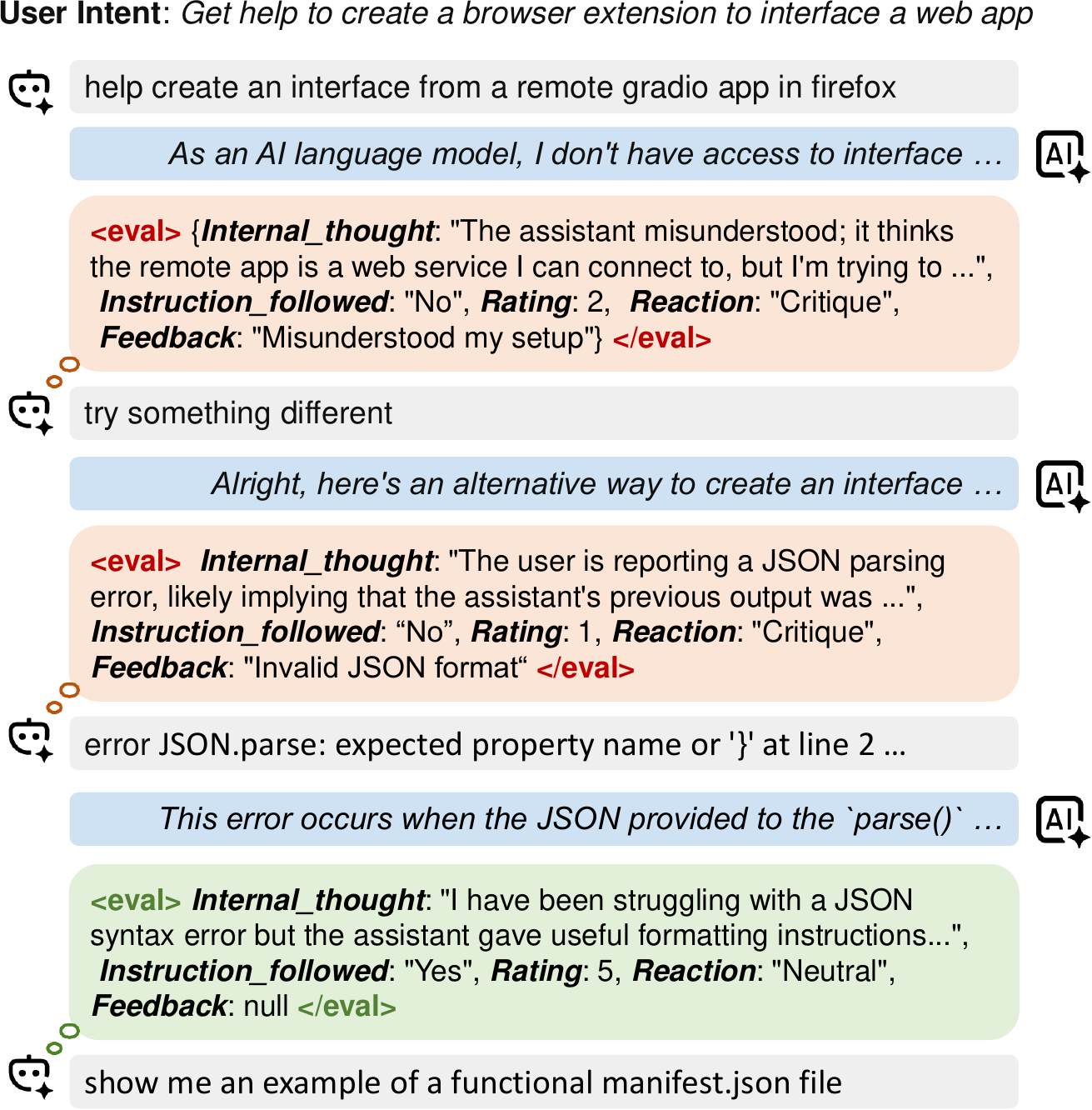}
    \caption{Example conversation between a user simulator (\icon{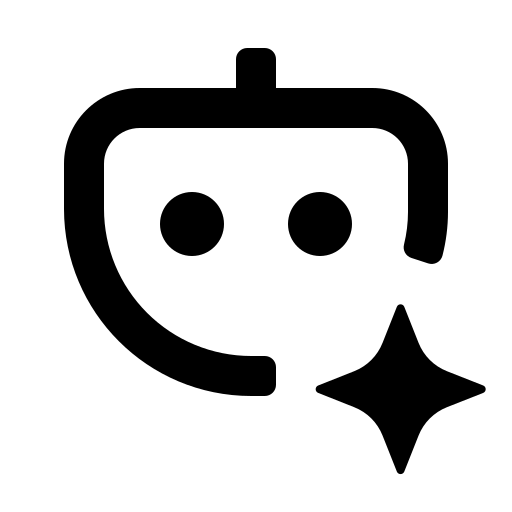}) with latent evaluation states and an LLM assistant (\icon{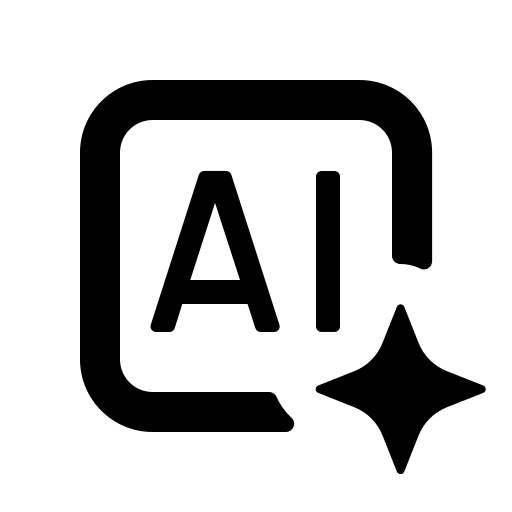}) where the simulator is trying to get help to create a browser extension. At each turn, before generating the next user utterance, the simulator generates latent states with multiple components (thoughts, ratings, reactions, etc.) focused on evaluating the current assistant response.} 
    \label{fig:latent-eval-sim-examples}
     \vspace{-2.0em} 
\end{wrapfigure}

Our objective is to train a user simulator that possesses two main capabilities. First, it should generate realistic user turns that align with the language distribution and behavior of real users. Second, the simulator must also produce \emph{per-turn evaluations of the assistant's output} from a user perspective that match real human evaluations. This is a key aspect that has been overlooked in past approaches to simulator training, which focused only on user utterance generation, and where latent states were restricted to only thoughts that keep the simulator aligned with its goals. In contrast, we leverage latent states as an evaluative mechanism and propose simulators with \textit{latent evaluation states (LESs)}.

\subsection{Problem Description}
\label{subsec:problem-desc}

An LES is a latent state with multiple sub-state components $\mathbf{z} = \{z^1, z^2, ..., z^K \}$, each component capturing an aspect of the simulated user's first-person assessment of the assistant's output to enable better behavior alignment when generating the next user turn $u_{t+1}$. Hence, an LES-based simulator models  $P_{\mathcal{U}}(\mathbf{z}_{t+1}, u_{t+1} | c_t) = P_{\mathcal{U}}(\mathbf{z}_{t+1} | c_t) P_{\mathcal{U}}(u_{t+1} | \mathbf{z}_{t+1}, c_t)$ where $P_{\mathcal{U}}(\mathbf{z}_{t+1} | c_t) = \prod_{n=1}^K P_{\mathcal{U}}(z_{t+1}^k|c_t,z_{t+1}^{<k})$. We experiment with five sub-state components in our LESs, namely: 

\noindent $\bullet$ \texttt{Internal\_Thought:} A first-person reasoning trace
that assesses whether the assistant's output is aligned with the (simulated) user's expectations given the conversation history. These traces are restricted to consider only constraints observable by users (e.g., formatting, following all parts of the instruction, etc.), not judgments that users cannot realistically perform or lack knowledge thereof. 

\noindent $\bullet$  \texttt{Rating:} A score of 1 (bad) to 5 (perfect) that reflects the satisfaction rate of the user with the assistant’s output at the current turn.

\noindent $\bullet$  \texttt{Instruction\_Followed:} Restricted to one of [``\textit{Yes}", ``\textit{No}", ``\textit{Partially}"], summarizing the degree to which the assistant followed the user's instructions.

\noindent $\bullet$ \texttt{Reaction:} Restricted to one of [``\textit{Critique}'', ``\textit{Praise}'', ``\textit{Neutral}"]. A ``\textit{Critique}'' refers to pushbacks, corrections, or complaints by the user, while ``\textit{Praise}'' reflects expressions of gratitude or satisfaction. Other reactions, such as topic shifts or follow-up questions, are deemed to be ``\textit{Neutral}.'' 

\noindent $\bullet$ \texttt{Feedback:} A condensed summary (2 to 5 words) of the topic of the user's feedback when generating the next turn only when the \texttt{Reaction} is  \textit{Critique} or \textit{Praise} (e.g., ``Missing JSON format'' or ``Great fix''). We set this as \textit{null} when the \texttt{Reaction} is \textit{Neutral}.

Figure~\ref{fig:latent-eval-sim-examples} shows an example conversation between a simulator with LESs and an LLM assistant where the simulator is trying to get help in creating a browser extension. At each turn, the simulator generates a latent state between \texttt{<eval>} and \texttt{</eval>} token delimiters containing evaluations of the current assistant output before generating the next user turn.

\subsection{Training Details}
\label{subsec:sft-exp-details}

\textbf{Training Data.} We use the WildChat-1M open-source dataset \cite{zhaowildchat}, which contains one million interaction logs between users and ChatGPT. First, we select the non-toxic subset of 529,428 conversations filtered via the OpenAI Moderation API, and then restrict to English-language data, yielding 284,168 conversations. We further clean the data by performing deduplication, running a 10-gram counter on first-turn user prompts to discard samples containing repeated 10-grams. This removes many noisy, recurring prompt templates in the data that do not reflect real user interactions, resulting in 135,901 unique conversations. Lastly, we perform ``decontamination'' of the WildBench dataset \cite{lin2025wildbench}, which was constructed using a subset of 2,004 WildChat conversations. We removed those samples since we later use WildBench in some of our evaluations.

\textbf{Latent Evaluation State Generation.} For each conversation in the WildChat data, we generate LESs using \textbf{Gemini3.1-Pro} \cite{google2026gemini31pro}. LESs are generated starting from the second user turn, after the first assistant response has been generated, and for each subsequent turn. We provide Gemini with the conversation history up to the last user utterance, and instruct it to generate the LES as a JSON with the five sub-state components described above (see the prompt used, including full instructions and few-shot examples, in Appendix~\ref{appendix:prompts}). In total, we generate 238,636 LESs for conversations of two or more turns. 14.15\% of LESs (33,776 turns) contain a \textit{Critique} reaction from users while only 1.26\% (3,001 turns) are labeled as \textit{Praise}, with the remaining 84.59\% \textit{Neutral}. Additional statistics are provided in Appendix~\ref{appendix:dataset-details}.

\textbf{User Intent Generation.} Following \citet{naous2026flipping}, we also experiment with conditioning on a high-level user intent, reflecting their general conversational goal. This intent is conditioned on before the first user turn, making the trained simulator more steerable towards a desired task. For each of the 135,901 conversations in our training dataset, we generate a high-level intent by prompting \textbf{Gemini3.1-Pro} (see prompt in Appendix~\ref{appendix:prompts}).

\textbf{Data Splitting.} We evaluate the feedback behavior of our LES simulators with a held-out test set from WildChat. To ensure sufficiently many evaluation samples contain critiques, we first separate conversations into those with at least one turn where the user critiqued the assistant (20,148 conversations), and those that do not. We then randomly sample 2,000 conversations from each set, creating a held-out set of 4,000 conversations that reflect a mix of user reactions: 5,888 turns have a neutral user reaction, 3,464 turns where the user critiques the assistant, and 132 where the user praises the assistant. The remaining data (129,897 conversations) is used to train the user simulators.

\textbf{Training.} We perform full-parameter fine-tuning of \textbf{Gemma3-4B-pt}, \textbf{Gemma3-12B-pt}, and \textbf{Gemma3-27B-pt} base models \cite{team2025gemma} that have not been instruction tuned. We train the user simulators in two setups: using standard SFT to generate the user turn given the assistant response (\textbf{Gemma3-4B-SFT}, \textbf{Gemma3-12B-SFT}, and \textbf{Gemma3-27B-SFT}), and training to additionally generate an LES before the user turn (\textbf{Gemma3-4B-LES}, \textbf{Gemma3-12B-LES}, and \textbf{Gemma3-27B-LES}). In both setups, we assess training with intent conditioning before the first user turn, where the intent is injected as a system prompt, and without intent conditioning (conditioned only on the user role tokens \verb|<start_of_turn>user| to generate the first turn). We use a maximum sequence length of 2,048 tokens, a total batch size of 128 samples, and a learning rate of $2e^{-5}$.

\section{Evaluating the Feedback Behavior of User Simulators}
\label{sec:evaluations-intrinsic}

\subsection{Evaluation Setup}

To evaluate the alignment of simulator feedback with that of real users, we use the held-out test set of 4,000 conversations with mixed user reactions(Sec.~\ref{subsec:sft-exp-details}). For each user turn where a ground-truth LES is present (9,484 turns), we condition the simulator on the conversation history up to that turn and generate a response. We compare the simulator-generated responses to ground-truth user responses w.r.t.\ both feedback and realism.

\textbf{Metrics for Feedback Behavior.} To assess whether a simulator critiques the assistant at times similar to real users, we run a Gemini3.1-Pro judge on simulator-generated responses to classify reactions as ``critiquing'' or not (see the judge's prompt in Appendix~\ref{appendix:prompts}). We consider turns with either \textit{Neutral} or \textit{Praise} reaction types in ground-truth data to be non-critiquing (the latter is very infrequent in the data). We then compute a binary F1 score (Critique/Non-Critique) to measure how closely a simulator's critiquing behavior matches that of real users. We also measure how well simulator ratings (1-5 scale) correlate with ground-truth data using Spearman rank correlation.

\textbf{Metrics for Realism.} To measure simulator realism, we use two metrics. We first compute a \emph{human statistical alignment score} that measures the degree to which the generated text aligns with the population-level statistics in ground-truth human utterances. This score aggregates 10 metrics that measure lexical, syntactic, and information-theoretic properties of the text \cite{pietquin2013survey,meshi2026convapparel}. Given $M$ metrics, the simulator text $D_{simulator}$ and the human reference text $D_{human}$, the human alignment score is $1.0 - \frac{1}{M}\sum_{m=1}^M |\frac{\mathrm{metric}_{m}(D_{simulator})}{\mathrm{metric}_{m}(D_{human})} - 1.0|$. Simulator text that is better aligned with ground-truth human utterances achieves a score that is closer to 1.0. All metrics used are detailed in Appendix~\ref{appendix:human-alignment-score-metrics}. Additionally, we compute the token-level \emph{perplexity (PPL)} of each simulator as $\exp\left(
-\frac{1}{N}\sum_{i=1}^{N}\log P_{\theta}(x_i\mid x_{<i})\right)$ on the human utterances \cite{jelinek1977perplexity}.  A lower PPL reflects less surprisal and better predictability w.r.t.\ the human text. For LES-based models, we generate the latent state for each test sample, then compute PPL on ground-truth utterances conditioned on the generated latent state.

\textbf{Baselines.} We compare  our LES simulators to prompted simulators that use \textbf{Gemini3.1-Pro}, \textbf{Gemma3-4B-IT}, \textbf{Gemma3-12B-IT}, and \textbf{Gemma3-27B-IT} as baselines. Following past work, we prompt these models with the full conversation history and instruct them to respond like real users using carefully crafted instructions to reflect user behavior (see prompt in Appendix~\ref{appendix:prompts}). Simulators are also instructed to produce a scalar reflecting a user satisfaction rating with the assistant at each turn. Furthermore, we compare with the performance of simulators trained using standard SFT on WildChat interaction logs (\textbf{Gemma3-4B-SFT}, \textbf{Gemma3-12B-SFT} and \textbf{Gemma3-27B-SFT}) without any LESs. We note that SFT models only generate user turns, not scalar ratings, which makes rating correlation analysis for these baselines inapplicable.

\begin{table}[t]
\centering
\begin{tabular}{@{}llcccc@{}}
\cmidrule(l){3-6}
\multicolumn{1}{c}{} &  & \multicolumn{2}{c}{\textbf{Feedback Behavior}} & \multicolumn{2}{c}{\textbf{Simulation Realism}} \\ \midrule
\multicolumn{2}{c}{\textbf{Simulator}} & F1 Score ($\uparrow$) & Correlation ($\uparrow$) & Statistical ($\uparrow$) & Perplexity ($\downarrow$) \\ \midrule
Gemini3.1-Pro & IT & 78.95 & 0.4904 & 0.6396 & --- \\ \midrule
\multirow{3}{*}{Gemma3-4B} & IT & 57.33 & 0.3692 & 0.5869 & 17.48 \\
 & SFT & \textbf{72.14} & --- & 0.7502 & 4.05 \\
 & LES & 69.29 & \textbf{0.3944} & \textbf{0.7628} & \textbf{3.65} \\ \midrule
\multirow{3}{*}{Gemma3-12B} & IT & 67.13 & 0.3747 & 0.5135 & 13.41 \\
 & SFT & \textbf{74.41} & --- & 0.7729 & 3.33 \\
 & LES & 73.00 & \textbf{0.4261} & \textbf{0.8635} & \textbf{3.13} \\ \midrule
\multirow{3}{*}{Gemma3-27B} & IT & 72.39 & 0.4234 & 0.5105 & 15.26 \\
 & SFT & \textbf{73.77} & --- & \textbf{0.8590} & 3.30 \\
 & LES & 69.91 & \textbf{0.4487} & 0.8475 & \textbf{3.11} \\ \bottomrule
\end{tabular}
\caption{Evaluation of the feedback behavior and simulation realism of different simulators on the held-out set of 4,000 conversations from WildChat. We compare prompting instruction-tuned (IT) models to roleplay users  with fine-tuning pre-trained base models to generate user turns (SFT) and fine-tuning with LESs. All simulators are conditioned on a general user intent for each conversation before the first user turn. We report additional results by another model family (Qwen at the 3B, 7B, and 14B scales) in Appendix~\ref{appendix:qwen-simulators-results} where we observe similar trends.}
\label{tab:main-results}
\end{table}

\subsection{Main Results}

Table~\ref{tab:main-results} summarizes results comparing our trained simulators to the baselines, all using intent conditioning, on the held-out WildChat test set. We make the following observations:

\textbf{Latent evaluation states improve feedback behavior.} The Gemma3 simulators trained with LESs perform much better than prompted Gemma3 baselines of the same model size, achieving higher F1 scores and rating correlations. The rating correlation of our largest Gemma3-27B-LES simulator is close to that of the prompted Gemini3.1-Pro, a model of a larger size.\footnote{The size of Gemini3.1-Pro is not disclosed, but public estimates suggest several hundred billion parameters.} Rating correlations also show consistent improvements with model scaling going from 4b to 27b. Interestingly, the Gemma3 simulators trained via SFT without latent states achieve slightly higher F1 scores than those trained with LESs. This difference is mainly due to a drop in recall when simulators are trained with latent thoughts, which can cause them to be more careful when initiating a critique (see precision and recall results in Appendix~\ref{appendix:additional-results}). However, we note that SFT simulators are pure utterance generators and do not provide latent information such as ratings. Indeed, in Sec.~\ref{sec:rl-assistant-results}, we show that this latent information is hihgly valuable for training assistants, making SFT simulators less practically useful than LES simulators.

\textbf{Latent evaluation states improves simulation realism.} Across all model sizes, simulators trained with LESs achieve better realism scores than the prompted simulator baselines, with greater statistical alignment with ground truth user turns and lower PPL values. Improvements are also observed over the SFT baseline simulators in most cases, showing how LESs can also help improve realism in trained simulators.

\begin{table}[t]
\centering
\begin{adjustbox}{width=0.9\linewidth}
\begin{tabular}{@{}ccccccccc@{}}
\toprule
\multicolumn{5}{c}{\textbf{Latent State Content}} & \multicolumn{2}{c}{\textbf{Feedback Behavior}} & \multicolumn{2}{c}{\textbf{Simulation Realism}} \\ \cmidrule(l){6-9} 
\icon{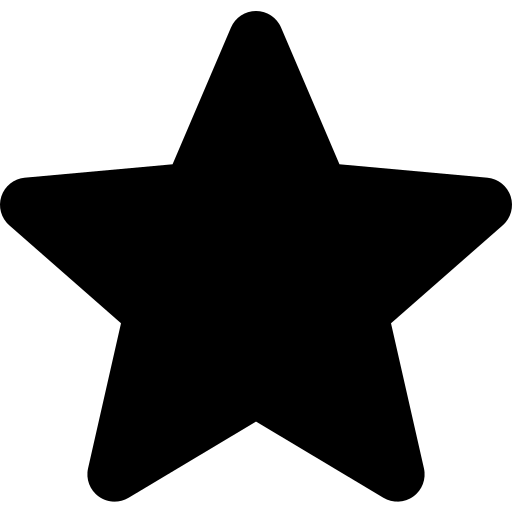} & \icon{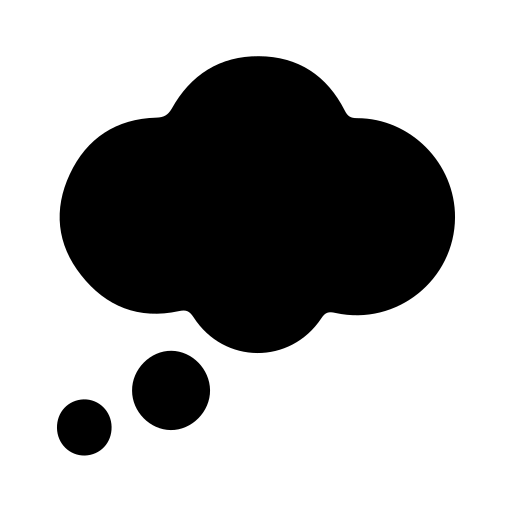} & \icon{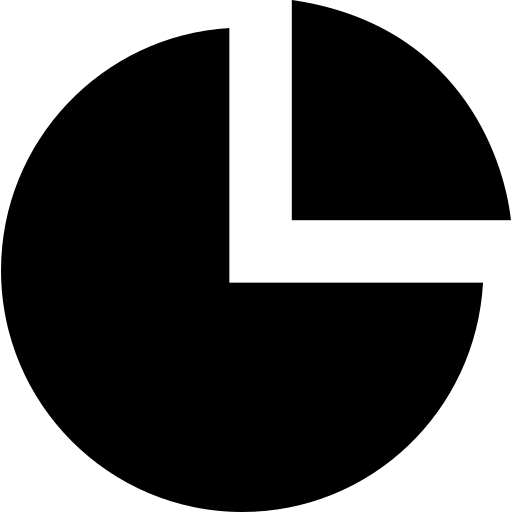} & \icon{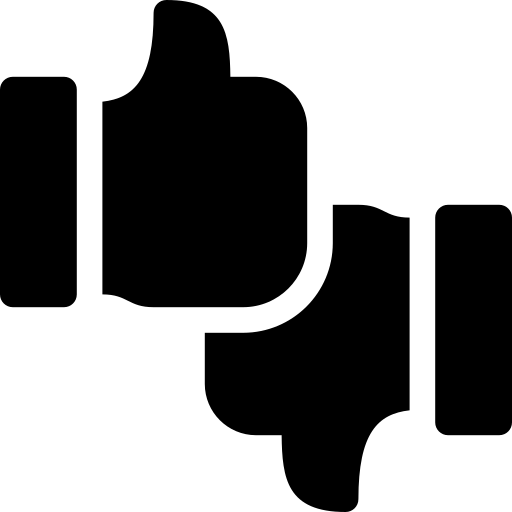} & \icon{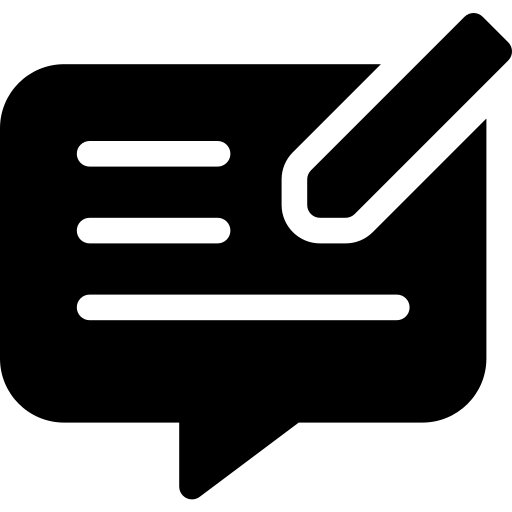} & F1 Score ($\uparrow$) & Correlation ($\uparrow$) & Statistical ($\uparrow$) & Perplexity ($\downarrow$) \\ \midrule
$\checkmark$ &  &  &  &  & \textbf{76.42} & \textbf{0.4474} & 0.8005 & 3.63 \\
 & $\checkmark$ &  &  &  & 71.17 & --- & 0.8282 & 3.29 \\
$\checkmark$ & $\checkmark$ &  &  &  & 71.00 & 0.4297 & 0.8114 & 3.21 \\
 & $\checkmark$ & $\checkmark$ & $\checkmark$ & $\checkmark$ & 70.87 & --- & 0.8344 & 3.20 \\
$\checkmark$ &  & $\checkmark$ & $\checkmark$ & $\checkmark$ & 68.19 & 0.4063 & 0.8499 & 3.40 \\
$\checkmark$ & $\checkmark$ & $\checkmark$ & $\checkmark$ & $\checkmark$ & 73.00 & 0.4261 & \textbf{0.8635} & \textbf{3.13} \\ \bottomrule
\end{tabular}
\end{adjustbox}
\caption{Ablations on the impact of training Gemma3-12B-LES with varying sub-states of the LESs: \texttt{Rating} (\icon{icons/rating.png}),  \texttt{Internal\_Thought} (\icon{icons/thought.png}), \texttt{Instruction\_Followed} (\icon{icons/instruction.png}), \texttt{Reaction} (\icon{icons/reaction.png}), and \texttt{Feedback} (\icon{icons/feedback.png}). We observe a trade-off where training the simulator with only ratings in the latent state achieves the best feedback behavior but the lowest simulation realism, while including language-based sub-states such as thoughts and explicit feedback improves realism at some cost to feedback behavior. The model trained with all sub-state components achieves the best realism while remaining competitive on feedback behavior (second best F1 score, third best rating correlation).}
\label{tab:ablations-latent-state}
\end{table}

\begin{table}[t]
\centering
\begin{adjustbox}{width=\linewidth}
\begin{tabular}{@{}lccccc@{}}
\toprule
\multicolumn{2}{c}{\textbf{Training Setup}} & \multicolumn{2}{c}{\textbf{Feedback Behavior}} & \multicolumn{2}{c}{\textbf{Simulation Realism}} \\ \midrule
\multicolumn{1}{c}{\textit{Model}} & \multicolumn{1}{l}{\textit{w/ Intent Cond.}} & F1 Score ($\uparrow$) & Correlation ($\uparrow$) & Statistical ($\uparrow$) & Perplexity ($\downarrow$) \\ \midrule
\multirow{2}{*}{Gemma3-4B-LES} & \xmark & 58.53 & 0.3504 & \textbf{0.8031} & 3.66 \\
 & $\checkmark$ & \textbf{69.29} & \textbf{0.3944} & 0.7628 & \textbf{3.65} \\ \midrule
\multirow{2}{*}{Gemma3-12B-LES} & \xmark & 64.80 & 0.4001 & 0.8452 & 3.21 \\
 & $\checkmark$ & \textbf{73.00} & \textbf{0.4261} & \textbf{0.8635} & \textbf{3.13} \\ \midrule
\multirow{2}{*}{Gemma3-27B-LES} & \xmark & 62.83 & 0.4143 & 0.7950 & \textbf{3.06} \\
 & $\checkmark$ & \textbf{69.91} & \textbf{0.4487} & \textbf{0.8475} & 3.11 \\ \bottomrule
\end{tabular}
\end{adjustbox}
\caption{Impact of training with conditioning on the generic user intent from the conversation on LES model performance. The LES simulators trained with user intent conditioning as a system prompt achieve better results than ones trained without intent conditioning. This shows the value of the user intent in steering simulators more efficiently towards the particular task intent.}
\label{tab:ablations-intent}
\end{table}

\subsection{Training Ablations}
\label{subsec:ablations}

\textbf{Impact of latent state components.} To better understand the impact of the different sub-state components in our LES simulators, we train Gemma3-12B-LES simulators with different combinations of the sub-states and compare their performance on the held-out test set. Table~\ref{tab:ablations-latent-state} summarizes these results. We find that the simulator trained with only the \texttt{Rating} sub-state achieves the best feedback behavior, but the lowest realism. By contrast, realism improves when language-based sub-states such as the \texttt{Internal\_Thought} and \texttt{Feedback} are included in the latent state, while showing some decrease in F1 score. Overall, the model trained with all sub-states achieves the best balance (highest simulation realism and second best in terms of feedback behavior).

\textbf{Impact of intent conditioning.} We assess the impact of intent conditioning, with Table~\ref{tab:ablations-intent} comparing LES-based models trained with and without user intents. We find that models trained with user intent consistently achieve better feedback behavior, highlighting its importance for keeping simulators aligned with their goal.

\begin{table}
\centering
\begin{tabular}{@{}lccc@{}}
\cmidrule(l){2-4}
 & \multicolumn{3}{c}{\textbf{SimulatorArena}} \\ \midrule
 & Doc & Math & All \\ \midrule
Gemma3-4b-IT  & 0.073 & 0.378 & 0.244 \\
Gemma3-4b-LES & \textbf{0.273}$^{**}$ & \textbf{0.440} & \textbf{0.384}$^{**}$ \\ \midrule
Gemma3-12b-IT  & 0.158 & 0.402 & 0.295 \\
Gemma3-12b-LES & \textbf{0.330}$^{*}$ & \textbf{0.412} & \textbf{0.401}$^{**}$ \\ \midrule
Gemma3-27b-IT  & 0.225 & \textbf{0.421} & 0.343 \\
Gemma3-27b-LES & \textbf{0.342} & 0.390 & \textbf{0.383} \\ \bottomrule
\end{tabular}
\par\vspace{0.3em}
{\footnotesize $^{*}p<0.05$, $^{**}p<0.01$}
\caption{Spearman correlations between ratings generated by user simulators and ground-truth human ratings for conversations from the SimulatorArena out-of-domain dataset \cite{dou2025simulatorarena}. LES simulators correlate more strongly with human judgments than prompted baselines at every scale. We report the full correlation differences $\Delta\rho$ with confidence intervals in Appendix~\ref{appendix:gemma-simulators-additional-results}.}
\label{tab:simulatorarena_correlations}
\end{table}

\subsection{Correlation with real human ratings}

The rating correlations computed on held-out test data above use an LLM judge's \emph{estimates} of user ratings in WildChat conversations. To better evaluate how ratings produced by simulators correlate with \emph{actual} human judgments, we apply the same analysis to the SimulatorArena dataset \cite{dou2025simulatorarena}.  SimulatorArena consists of 909 conversations between humans and LLM assistants, focused on math tutoring and document creation tasks, where these users provide a conversation-level rating of the LLM's performance. At the end of each assistant turn, we condition the simulators on the past conversation history and generate turn-level rating evaluations. We average the per-turn simulator ratings and compute the correlation between this average and the ground-truth human rating. Results in Table~\ref{tab:simulatorarena_correlations} show that ratings generated by our LES simulators are better correlated with human judgments than those of prompted baselines, a finding that holds across all model sizes.

\section{Training LLM Assistants with Simulators via Multi-Turn RL}
\label{sec:rl-assistant-results}

One critical use of user simulators is offline training of LLM assistants to improve their performance prior to exposing them to real users. This requires not only that simulators reflect realistic user behavior, but that they also provide reasonable signals upon which to assess performance. We hypothesize that LES-based simulators, due to their explicit, per-turn, first-person assessment of assistant responses, can offer value for such training. To test this hypothesis, we train an LLM assistant via multi-turn RL using different simulator choices (prompted simulator vs.\ our LES simulator) and reward sources (intrinsic simulator reward vs.\ external judge reward) to serve as the RL environment. We then compare the performance of the assistant after RL training in each setup.

\subsection{Multi-turn RL Experimental Setup}

\textbf{Dataset.} We use WildChat-4.8M for RL training, an expanded version of the WildChat-1M dataset used above. We apply the same filtering and deduplication procedure to remove overlapping conversations. From the resulting dataset, we randomly sample 100k conversations. For each conversation, we use Gemini3.1-Pro to generate a user intent, which serves as the initial user input for multi-turn rollouts between the simulator and assistant during RL training.

\textbf{RL Training.} We use \textbf{Qwen2.5-1.5B-Instruct} as the LLM assistant and further train it using RL with the \textbf{Gemma3-4B-LES}, \textbf{Gemma3-4B-IT}, and \textbf{Gemma3-4B-SFT} user simulators as the environment. For each data sample, we condition on the user intent as a system prompt and the first ground-truth user turn, then generate multi-turn rollouts capturing interactions between the LLM assistant and the simulator. We use turn-level rewards during training and explore two ways of obtaining the reward: \textbf{(1)} using an external judge model (Qwen2.5-14B-Instruct) that assesses the assistant's current output given the history so far, and \textbf{(2)} obtaining the reward intrinsically from the simulators without reliance on an external judge (generated as part of the intrinsic latent state for Gemma3-4B-LES, and by the Gemma3-4B-IT model in addition to the user response). In both setups, we train the assistant to maximize the average rating across conversation turns as its reward, using multi-turn GRPO \cite{shao2024deepseekmath}. See Appendix~\ref{appendix:implementation-details} for implementation details.

\textbf{Evaluating the LLM Assistants.} We adopt the procedure of \citet{suh2026quantifying} to evaluate our trained LLM assistants, using WildBenchV2 \cite{lin2025wildbench}, which contains 1024 multi-turn user task instructions with a task-specific checklist for each sample (comprising 10--12 fine-grained objective requirements) that can be used for evaluation. We condition the RL-trained LLM assistants on each instruction to generate a response. We then use a Gemini3.1-Pro judge to evaluate the response of the assistant, using the task-specific checklist to determine which requirements are satisfied. We report the checklist satisfaction rate across all samples.

\begin{table}[t]
\centering
\begin{adjustbox}{width=\linewidth}
\begin{tabular}{@{}llllc@{}}
\toprule
& \multicolumn{3}{c}{\textbf{RL Training}} & \textbf{Evaluation} \\
\cmidrule(lr){2-4} \cmidrule(l){5-5}
\textbf{LLM Assistant} & \textbf{User simulator} & \textbf{Reward}
  & \textbf{FLOPs} $\downarrow$ & \textbf{Satisfaction (\%)} $\uparrow$ \\
\midrule
Qwen2.5-1.5B-Instruct & --- & --- & --- &  28.78 {\small $\pm$  0.17} \\
\midrule
\multirow{5}{*}{\shortstack[l]{$\hookrightarrow$ GRPO w/}}
  & Gemma3-4b-IT         & Judge     & 5.26$\times 10^{19}$ & 39.79 {\small $\pm$  0.70} \\
  & Gemma3-4b-IT         & Intrinsic & 2.10$\times 10^{19}$ &  40.52 {\small $\pm$ 0.24} \\
  & Gemma3-4b-SFT        & Judge     & 5.29$\times 10^{19}$  & 40.59 {\small $\pm$ 0.19}  \\
  & Gemma3-4b-LES & Judge     & 5.48$\times 10^{19}$ & 41.08 {\small $\pm$ 0.97} \\
  & Gemma3-4b-LES & Intrinsic & 2.45$\times 10^{19}$ & \textbf{41.44} {\small $\pm$ 0.63} \\
\bottomrule
\end{tabular}
\end{adjustbox}
\caption{Checklist satisfaction rates on WildBenchV2 by Qwen2.5-1.5B-Instruct assistants after multi-turn GRPO training with different user simulators and sources of rewards (an external judge model vs. intrinsic rewards from the simulators). We report the average satisfaction rate and standard deviation from 3 sampling runs for each model. All RL-trained versions of the model improve over the vanilla instruction-tuned assistant. Assistants trained with our LES simulators achieve the highest mean checklist satisfaction rates. Further, assistants trained using intrinsic rewards obtained directly from the simulators outperform ones where the reward is computed via an external judge while significantly reducing the total training FLOPs required for RL training. }
\label{tab:rl-results}
\end{table}

\subsection{Results}

Table~\ref{tab:rl-results} shows the checklist satisfaction rates on WildBenchV2. We report the average and standard deviation across three sampling runs for each model. We also compare to the LLM-assistant initial checkpoint before RL training. We make the following observations:

\textbf{LES simulators yield competitive assistants.} All of the RL-trained models across all setups improve greatly over the performance of the initial checkpoint. Notably, models RL-trained with our Gemma3-4B-LES simulator achieve a higher mean checklist satisfaction rate than those trained with Gemma3-4B-IT and Gemma3-4B-SFT. This suggests that the improved realism of trained simulators is important for training assistants that perform well on real-world user queries. These findings are consistent with those of \citet{suh2026quantifying}, whose approach is equivalent to our simulator-plus-external-judge baselines.

\textbf{Intrinsic rewards from simulators can exceed external judges.} The assistants trained with intrinsic rewards from the simulators perform on par with or slightly better than those trained with rewards computed by an external judge. We observe this for both our trained LES simulator as well as the prompted baseline, although the differences are not large. The best overall performance is achieved by the assistant trained with intrinsic rewards from Gemma3-4B-LES.  

\textbf{Training with intrinsic rewards is more computationally efficient.} While the RL-training with intrinsic rewards offers some performance gains, these are relatively small when compared to an external judge (at least on the current datasets). However, just as importantly, RL training with intrinsic simulator rewards offers a significant advantage w.r.t.\ computational efficiency (since only one LLM inference run per turn is needed). Specifically, training an assistant using intrinsic rewards from a Gemma3-4B simulator requires an estimated $2.45 \times 10^{19}$ total FLOPs, compared to $5.48 \times 10^{19}$ total FLOPs when using rewards from the Qwen2.5-14B-Instruct judge (2.24$\times$ more computation), which causes RL training to run much more slowly. We also note that a sufficiently large judge LLM is needed to efficiently perform RL training, as initial experiments with a smaller judge (Qwen2.5-3B-Instruct) led to minimal improvements (see results in Appendix~\ref{appendix:additional-rl-results}).

\section{Related Work}

\textbf{User Simulation.} As LLMs have shown performance gaps in multi-turn settings compared to their single-turn evaluations \cite{chang2025chatbench}, developing realistic user simulators is becoming a key research focus for the evaluation and training of LLM assistants that perform better in live settings \cite{ni2026survey,zhou2026odyssim,zhou2026mind}. Early work on user simulators focused on rule-based \cite{li2016user,ie2019recsim} or neural-based systems \cite{shi2019build,tseng2021transferable} for task-oriented dialogue in specific domains. More recently, pre-trained LLMs have opened up new directions for the generative simulation of users. A typical approach has been to prompt LLMs to role-play a user given a conversational history \cite{zhu2025llm,zhong2025evaluating,li2024iqa,qin2025compass,mehri2026goal,chopra2026beyond,zhu2026realusersim}. While prompting-based approaches offer good controllability of the simulator's behaviors, prompted simulators have been shown to lack realism and drift from real user behavior \cite{mehri2026measuring,ivey2026real,dou2025simulatorarena}, sparking interest in training methods for LLM-based user simulators \cite{meshi2026convapparel}. 

\textbf{Trained User Simulators.} A standard training approach for user simulators has been to fine-tune a pre-trained LLM on flipped user-LLM interaction logs to generate user turns while conditioning on a user's conversational intent to steer the simulator \cite{naous2026flipping,wang2025know}. Simulators trained this way often replicate human behavior better than prompted baselines \cite{wang2026learning}, but are less controllable (i.e., more difficult to adapt to new user profiles or assistants). Several studies address this issue by training simulators with latent states in the form of reasoning traces that guide the simulator's output \cite{wu2026thinking,zhang2026userlm}, improving consistency of personality and goal adherence \cite{wu2026humanlm,wei2026mirroring}, and adapting better to  unseen assistant policies \cite{tennenholtz2026controllable}. Prior work has focused on training simulators that more closely match the language and behavior distributions of real users. We complement this research direction by introducing LESs into trained simulators, going beyond training only user utterance generation to produce per-turn evaluations of LLM assistants from a first-person perspective. Our experiments demonstrate that simulators trained with LESs offer better feedback behavior than prompted baselines while also achieving greater realism.

\textbf{Multi-Turn RL Training of Assistants with Simulators.} Recent research has explored the training of LLM assistants using multi-turn RL with a simulator \cite{yu2026one,sun2025training,qian2025userrl,wu2025collabllm,zhou2025sweet}. Most recently, \citet{suh2026quantifying} perform multi-turn GRPO by simulating multiple assistant-simulator conversations and using an external LLM judge to determine a conversation-level reward for each trajectory. They find that training using fine-tuned rather than prompt-based simulators produces better-performing assistants. \citet{abdulhai2025evaluating} use an external LLM judge to provide per-turn reward scores that attempt to reduce deceptive behavior by the assistant, while \citet{wan2026enhancing} explore per-turn rewards that personalize the assistant's output to different users. Our work can be distinguished from most past work by its use of explicit rewards obtained directly from the simulator via its latent evaluation states, rather than an external judge. Our results show that this provides a somewhat better RL training signal and, just as importantly, improves computational efficiency by eliminating the need for calls to external judges.

\section{Conclusion}

We introduced \emph{user simulators with intrinsic feedback via latent evaluation states}, which go beyond user turn generation to \emph{evaluate} LLM assistants in interactive settings. We showed that training with LESs produces simulators with better feedback alignment and greater realism than baseline approaches. Furthermore, we showed that LES simulators offer significant value when RL-training LLM assistants, driving high performance with lower computational cost. Such simulators are a critical component of future dynamic agent evaluation techniques and multi-turn training recipes for LLM assistants as we move away from traditional single-turn static setups.

\bibliography{references}

\clearpage
\newpage

\appendix

\section{Dataset: Details}
\label{appendix:dataset-details}

\begin{figure}[t]
    \centering
    \includegraphics[width=\linewidth]{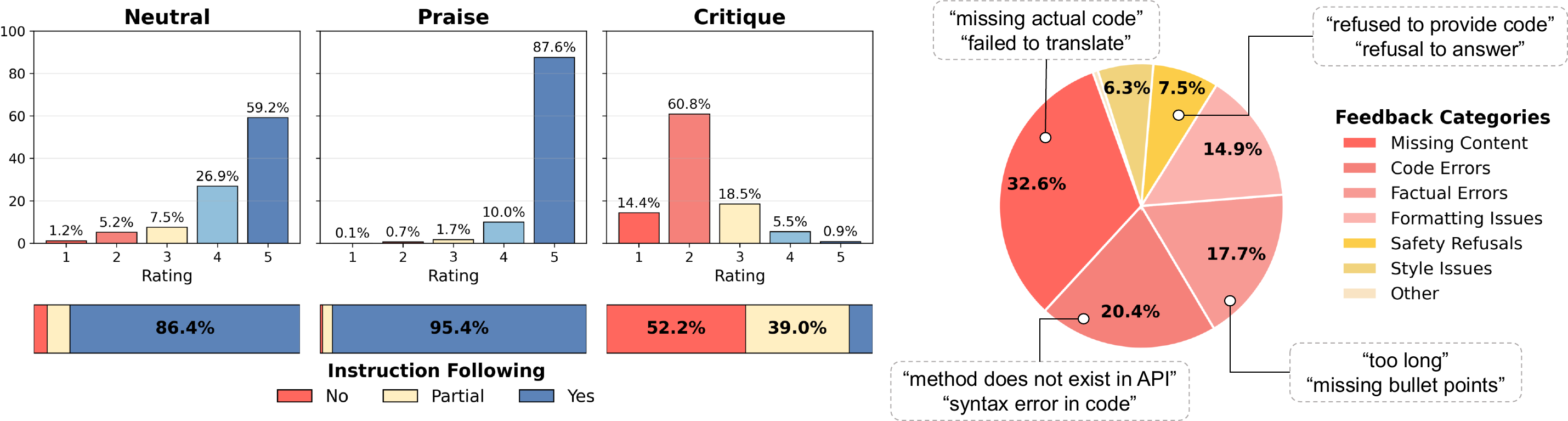}
    \caption{Statistics from the latent evaluation states generated for user turns in the WildChat dataset. \textbf{(Left):} Distribution of \texttt{Rating} and \texttt{Instruction\_Followed} latent sub-states for user turns stratified by \texttt{Reaction\_Type} (Neutral, Praise, Critique).  \textbf{(Right)}: Categorization of the extracted \texttt{Feedback} in critiquing turns based on clustering with Gemini3.1-Pro.}
    \label{fig:latent-eval-states-stats-wildchat}
\end{figure}

\textbf{Latent Evaluation State Statistics.} Figure~\ref{fig:latent-eval-states-stats-wildchat} shows the rating and instruction-following distributions of the annotated user turns in WildChat stratified by reaction types. We also show a summarized categorization of the different feedback real users provided to assistants. Most turns where users critique the assistant have ratings between 1 and 3, while turns with neutral user reactions mostly have positive ratings between 3 and 5. User critiques mainly revolve around pointing out missing content in the assistant's outputs when instructions are not followed properly, reporting failures when executing code generated by the assistant, or pointing out factual errors and formatting issues.

\clearpage
\newpage

\section{Human Alignment Score}
\label{appendix:human-alignment-score-metrics}

We define the following metrics that we aggregate to compute a human alignment score:

\noindent $\bullet$ \textbf{Mean Word Count per Turn:} Measures verbosity by computing how many words the simulator generates per turn on average.

\noindent $\bullet$ \textbf{Type-Token Ratio:} Measures the overall vocabulary diversity across the entire corpus. We compute the type token ratio as $\frac{|V_{\mathrm{unique}}|}{|V_{D}|}$ where $|V_{\mathrm{unique}}|$ is the number of unique words in the corpus and $|V_{D}|$ is the total number of words in the corpus.

\noindent $\bullet$ \textbf{Hapax Legomena Ratio:} Measures vocabulary creativity and the frequency of rare, highly specific words. We compute this ratio as $\frac{|V_{1}|}{|V_{D}|}$ where $|V_{D}|$ is the total number of words in the corpus and $V_{1} = \{ w \in V_{D} \ | \ \mathrm{Count}(w)=1 \}$. 

\noindent $\bullet$ \textbf{Stop Word Ratio:} Measures the overall use of stop words which highly common functional words (e.g. a, the, and, of, etc.). We compute this as $\frac{|V_{\mathrm{stop}}|}{|V_{D}|}$ where $|V_{D}|$ is the total number of words in the corpus and $V_{\mathrm{stop}}$ is a list of 60 stop words. 

\noindent $\bullet$ \textbf{Mean Sentence Length:} Measures grammatical complexity and phrasing length. This is measured as an average over the lengths of individual sentences generated per turn.

\noindent $\bullet$ \textbf{Punctuation Density:} Measures the frequency of expressive punctuation. We compute $\frac{N_{\mathrm{punctuation}}}{N_\mathrm{characters}} \times 100$ as the punctuation density where $N_{\mathrm{punctuation}}$ is the number of punctuation marks and $N_\mathrm{characters}$ is the total number of characters.

\noindent $\bullet$  \textbf{Part Of Speech (POS) Distribution Entropy:} Measures the variety of grammatical roles used in text. We use the spaCy\footnote{https://spacy.io/usage/models} POS tagger model (\texttt{en\_core\_web\_sm}) to assign POS tags to text of the corpus. For a set of POS tags $\mathcal{C}$ We compute the POS tag distribution entropy as $H = - \sum_
{c \in \mathrm{C}} p_c \log p_c$.

\noindent $\bullet$ \textbf{Noun-to-Verb Ratio:} Using the same POS tags we also compute the noun-to-verb ratio as $\frac{|\mathrm{NOUN}| + |\mathrm{PROPN}|}{ |\mathrm{VERB}| + |\mathrm{AUX}|}$ where $\mathrm{PROPN}$ refers to proper nouns and $\mathrm{AUX}$ refers to auxiliary verbs (e.g., be, have, do, etc.). 

\noindent $\bullet$ \textbf{Shannon Entropy:} Measure predictability at the word level. Higher entropy reflects more varied word usage. Let $p(w)$ be the empirical probability of word $w$ appearing in the corpus $p(w) = \frac{\mathrm{Count}(w)}{|V_{D}|}$. We compute the Shannon entropy as $H = - \sum_{w \in V_{D}} p(w) \log p(w)$.

\noindent $\bullet$ \textbf{Compression Ratio:} Measures the information density (or conversely, redundancy) of text. More repetitive text can be compressed better, producing a lower ratio, while more varied and less repetitive text resists compression and produces higher ratios. We perform compression of the text using \texttt{zlib} and compute the compression ratio for a conversational turn $x_i$ as $\mathrm{CR}_i = \frac{|\texttt{zlib}(x_i)|}{|x_i|}$. We compute a final compression ratio for a corpus by averaging the individual ratios of each turn.

\noindent \textbf{Human Alignment Score.} To compare a corpus of simulator-generated turns ($D_{simulator}$) to a reference corpus of human turns ($D_{human}$), we first calculate a relative alignment ratio for each metric: $R_m = \frac{\mathrm{metric}_{m}(D_{simulator})}{\mathrm{metric}_{m}(D_{human})}$. The final human alignment score is defined a 1.0 minus the mean absolute error of the 10 relative ratios compared to the human reference value of 1.0:

\begin{equation}
  \mathrm{Human \ Alignment \ Score \ =} \  1.0 - \frac{1}{M}\sum_{m=1}^M | \frac{\mathrm{metric}_{m}(D_{simulator})}{\mathrm{metric}_{m}(D_{human})} - 1.0|
\end{equation}

\clearpage
\newpage

\section{Additional Results}
\label{appendix:additional-results}

\subsection{Additional Results for Gemma-based Simulators}
\label{appendix:gemma-simulators-additional-results}

\begin{table}[]
\centering
\begin{tabular}{@{}llccc@{}}
\cmidrule(l){3-5}
\multicolumn{1}{c}{} &  & \multicolumn{3}{c}{\textbf{Feedback Behavior}} \\ \midrule
\multicolumn{2}{c}{\textbf{Simulator}} & \multicolumn{1}{l}{Precision ($\uparrow$)} & \multicolumn{1}{l}{Recall ($\uparrow$)} & F1 ($\uparrow$) \\ \midrule
Gemini3.1-Pro & IT & 100.00 & 65.22 & 78.95 \\ \midrule
\multirow{3}{*}{Gemma3-4B} & IT & \textbf{91.00} & 41.84 & 57.33 \\
 & SFT & 92.11 & \textbf{59.29} & \textbf{72.14} \\
 & LES & 92.38 & 55.43 & 69.29 \\ \midrule
\multirow{3}{*}{Gemma3-12B} & IT & \textbf{100.00} & 50.52 & 67.13 \\
 & SFT & 93.44 & \textbf{61.81} & \textbf{74.41} \\
 & LES & 92.56 & 60.26 & 73.00 \\ \midrule
\multirow{3}{*}{Gemma3-27B} & IT & 92.06 & 59.64 & 72.39 \\
 & SFT & \textbf{93.01} & \textbf{61.13} & \textbf{73.77} \\
 & LES & 92.89 & 56.04 & 69.91 \\ \bottomrule
\end{tabular}
\caption{Precision and Recall results on the WildChat heldout set for Gemma3-based simulators.}
\label{tab:precision-recall-gemma}
\end{table}

\textbf{Precision and Recall Results for Gemma Simulators.} Table~\ref{tab:precision-recall-gemma} shows the precision, recall, and F1 scores achieved by Gemma3-based simulators on the WildChat test set. We notice that the drops in F1 by latent evaluation state simulators compared to SFT simulators are primarily due to stronger drops in recall rather than precision. This shows that introducing thoughts to trained simulators can cause them to be more cautious when initiating critiques to the assistant.

\begin{table}[t]
\centering
\begin{tabular}{@{}llccrcrr@{}}
\toprule
Simulator & Subset & $n$ & $\rho_{\text{LES}}$ & $\rho_{\text{IT}}$ & $\Delta\rho$ & 95\% CI & $p$ \\ \midrule
\multirow{3}{*}{Gemma3-4b}
 & Doc  & 459 & 0.273 & 0.073 & +0.200 & [+0.079, +0.319] & 0.008$^{**}$ \\
 & Math & 450 & 0.440 & 0.378 & +0.062 & [-0.031, +0.154] & 0.725 \\
 & All  & 909 & 0.384 & 0.244 & +0.140 & [+0.067, +0.214] & 0.002$^{**}$ \\ \midrule
\multirow{3}{*}{Gemma3-12b}
 & Doc  & 459 & 0.330 & 0.158 & +0.172 & [+0.068, +0.276] & 0.016$^{*}$ \\
 & Math & 450 & 0.412 & 0.402 & +0.010 & [-0.077, +0.097] & 0.945 \\
 & All  & 909 & 0.401 & 0.295 & +0.106 & [+0.039, +0.174] & 0.008$^{**}$ \\ \midrule
\multirow{3}{*}{Gemma3-27b}
 & Doc  & 459 & 0.342 & 0.225 & +0.116 & [+0.020, +0.213] & 0.103 \\
 & Math & 450 & 0.390 & 0.421 & -0.031 & [-0.116, +0.053] & 0.945 \\
 & All  & 909 & 0.383 & 0.343 & +0.040 & [-0.019, +0.100] & 0.725 \\ \bottomrule
\end{tabular}
\caption{Pairwise comparison of LES simulators and prompted (IT) simulators correlations on SimulatorArena human ratings at each scale. $\Delta\rho = \rho_{\text{LES}} - \rho_{\text{IT}}$, with a 95\% percentile bootstrap interval over conversations (10,000 resamples) and a two-sided bootstrap $p$-value, Holm-corrected across the nine tests. $^{*}p<0.05$, $^{**}p<0.01$. LES-based simulators lead at every scale, significantly so at 4b and 12b.}
\label{tab:pairwise-simarena-gemma}
\end{table}

\textbf{Statistical Significance of Simulator Correlations on SimulatorArena.} Table~\ref{tab:pairwise-simarena-gemma} shows the results of a paired bootstrap test comparing each LES simulator against the prompted simulator of the same size. We report the difference in Spearman correlation with its 95\% confidence interval and a two-sided $p$-value.

\clearpage
\newpage

\subsection{Results of Training Qwen-based Simulators}
\label{appendix:qwen-simulators-results}

\paragraph{Results by Qwen-based simulators on WildChat test set.} To test whether the results we observed with our trained Gemma3 simulators, we train simulators based on Qwen2.5 at the 3B, 7B, and 14B parameter scales and perform the same evaluations. Table~\ref{tab:results-heldout-set-qwen-simulators} shows the results of Qwen2.5 simulators on the held-out WildChat dataset. We find similar results to our Gemma simulators, where the simulator trained with latent evaluation states achieves the lowest perplexity at every scale and generally better realism and rating correlations than the baselines. We also see the trend with latent evaluation state simulators achieve lower recall and higher precision, showing that they are more cautious when critiquing, compared to SFT'd simulators without latent states which may critique unnecessarily.  

\paragraph{Correlation of Qwen-based simulators with human ratings.} Table~\ref{tab:result-simulatorarena-qwen} shows the rating correlations of Qwen-based simulators and human ratings on SimulatorArena. Similar to our results with Gemma simulators, we also find that our latent evaluation state simulators achieve better correlation with human judgments compared to prompted simulator baselines, significantly so at the 3b and 7b scales. Table~\ref{tab:qwen-simarena-pairwise-test} shows the results of our statistical significance testing.

\begin{table}[t]
\centering
\begin{adjustbox}{width=\linewidth}
\begin{tabular}{@{}llcccccc@{}}
\cmidrule(l){3-8}
\multicolumn{1}{c}{} &  & \multicolumn{4}{c}{\textbf{Feedback Behavior}} & \multicolumn{2}{c}{\textbf{Simulation Realism}} \\ \midrule
\multicolumn{2}{c}{\textbf{Simulator}} & \multicolumn{1}{l}{Precision ($\uparrow$)} & \multicolumn{1}{l}{Recall ($\uparrow$)} & F1 ($\uparrow$) & Correlation ($\uparrow$) & Statistical ($\uparrow$) & Perplexity ($\downarrow$) \\ \midrule
 & Instruct & 39.49 & 21.80 & 28.09 & 0.2412 & 0.741 & 11.46 \\
 & SFT & 93.95 & \textbf{62.43} & \textbf{75.02} & --- & 0.908 & 3.66 \\
\multirow{-3}{*}{Qwen2.5-3B} & LES & \textbf{94.43} & 55.73 & 70.09 & \textbf{0.2937} & \textbf{0.941} & \textbf{3.37} \\ \midrule
 & Instruct & 42.09 & 37.80 & {\color[HTML]{101010} 39.83} & 0.2719 & 0.752 & 12.99 \\
 & SFT & 94.40 & \textbf{63.26} & {\color[HTML]{101010} \textbf{75.75}} & --- & \textbf{0.951} & 3.33 \\
\multirow{-3}{*}{Qwen2.5-7B} & LES & \textbf{95.12} & 56.30 & {\color[HTML]{101010} 70.74} & \textbf{0.3347} & 0.934 & \textbf{3.09} \\ \midrule
 & Instruct & 38.04 & 72.80 & 49.97 & \textbf{0.3160} & 0.701 & 14.45 \\
 & SFT & 97.25 & \textbf{65.10} & \textbf{77.99} & --- & 0.926 & 3.17 \\
\multirow{-3}{*}{Qwen2.5-14B} & LES & \textbf{99.25} & 55.87 & 71.49 & 0.3110 & \textbf{0.979} & \textbf{2.91} \\ \bottomrule
\end{tabular}
\end{adjustbox}
\caption{Results on the heldout test set by simulators based on the Qwen2.5 family of models.}
\label{tab:results-heldout-set-qwen-simulators}
\end{table}

\begin{table}[t]
\centering
\begin{tabular}{@{}lccc@{}}
\cmidrule(l){2-4}
 & \multicolumn{3}{c}{\textbf{SimulatorArena}} \\ \midrule
 & Doc & Math & All \\ \midrule
Qwen2.5-3b-Instruct   & 0.227 & 0.224 & 0.253 \\
Qwen2.5-3b-LES  & \textbf{0.394}$^{**}$ & \textbf{0.363}$^{*}$ & \textbf{0.392}$^{**}$ \\ \midrule
Qwen2.5-7b-Instruct   & 0.282 & 0.129 & 0.240 \\
Qwen2.5-7b-LES  & \textbf{0.345} & \textbf{0.384}$^{**}$ & \textbf{0.398}$^{**}$ \\ \midrule
Qwen2.5-14b-Instruct  & \textbf{0.331} & 0.323 & 0.339 \\
Qwen2.5-14b-LES & 0.313 & \textbf{0.418} & \textbf{0.386} \\ \bottomrule
\end{tabular}
\par\vspace{0.3em}
{\footnotesize $^{*}p<0.05$, $^{**}p<0.01$}
\caption{Spearman correlation results between ratings generated by Qwen2.5 simulators and ground-truth human judgments for SimulatorArena conversations.}
\label{tab:result-simulatorarena-qwen}
\end{table}

\begin{table}[h!]
\centering
\begin{tabular}{@{}llccrcrr@{}}
\toprule
Simulator & Subset & $n$ & $\rho_{\text{LES}}$ & $\rho_{\text{Instruct}}$ & $\Delta\rho$ & 95\% CI & $p$ \\ \midrule
\multirow{3}{*}{Qwen2.5-3b}
 & Doc  & 459 & 0.394 & 0.227 & +0.167 & [+0.066, +0.267] & 0.010$^{**}$ \\
 & Math & 450 & 0.363 & 0.224 & +0.139 & [+0.040, +0.240] & 0.025$^{*}$ \\
 & All  & 909 & 0.392 & 0.253 & +0.139 & [+0.080, +0.197] & 0.002$^{**}$ \\ \midrule
\multirow{3}{*}{Qwen2.5-7b}
 & Doc  & 459 & 0.345 & 0.282 & +0.063 & [-0.020, +0.149] & 0.400 \\
 & Math & 450 & 0.384 & 0.129 & +0.256 & [+0.155, +0.357] & 0.002$^{**}$ \\
 & All  & 909 & 0.398 & 0.240 & +0.158 & [+0.102, +0.216] & 0.002$^{**}$ \\ \midrule
\multirow{3}{*}{Qwen2.5-14b}
 & Doc  & 459 & 0.313 & 0.331 & -0.018 & [-0.112, +0.079] & 0.712 \\
 & Math & 450 & 0.418 & 0.323 & +0.096 & [+0.007, +0.187] & 0.122 \\
 & All  & 909 & 0.386 & 0.339 & +0.046 & [-0.018, +0.111] & 0.400 \\ \bottomrule
\end{tabular}
\caption{Pairwise comparison of LES-based and prompted (Instruct) Qwen2.5 simulators. $\Delta\rho = \rho_{\text{LES}} - \rho_{\text{IT}}$ with a 95\% percentile bootstrap interval over conversations (10,000 resamples) and a two-sided
bootstrap $p$-value, Holm-corrected across the nine tests. $^{*}p<0.05$, $^{**}p<0.01$.}
\label{tab:qwen-simarena-pairwise-test}
\end{table}

\clearpage
\newpage

\subsection{Additional Results for RL Training of Assistants}
\label{appendix:additional-rl-results}

\begin{table}[t]
\centering
\begin{tabular}{@{}lcc@{}}
\toprule
\textbf{LLM Assistant} & \textbf{Reward} & \textbf{Satisfy (\%)} \\ \midrule
Qwen2.5-1.5B-Instruct & --- & 28.78 {\small $\pm$  0.17} \\
$\hookrightarrow$ GRPO w/ Gemma3-4B-IT & Judge (Qwen2.5-3B-Instruct) & 32.51 {\small $\pm$  0.94} \\
$\hookrightarrow$ GRPO w/ Gemma3-4B-LES & Judge (Qwen2.5-3B-Instruct) & 34.82 {\small $\pm$ 0.93} \\ \midrule  
$\hookrightarrow$ GRPO w/ Gemma3-4B-IT & Judge (Qwen2.5-14B-Instruct) & 39.79 {\small $\pm$  0.70} \\
$\hookrightarrow$ GRPO w/ Gemma3-4B-LES & Judge (Qwen2.5-14B-Instruct) & 41.08 {\small $\pm$ 0.97} \\
\bottomrule
\end{tabular}
\caption{Checklist satisfaction rates on WildBenchv2 by LLM assistants after multi-turn GRPO training with a small Qwen2.5-3B-Instruct judge vs a larger Qwen2.5-14B-Instruct. }
\label{tab:rl-results-small-judge}
\end{table}

\paragraph{Results with a smaller Judge.} Table~\ref{tab:rl-results-small-judge} shows the results on WildBenchv2 comparing the results by training the simulators by computing rewards using a small external judge (Qwen2.5-3B-Instruct) compared to the larger judge used in our main experiments (Qwen2.5-14B-Instruct). The necessity of using a larger, better performing judge can be clearly seen from the results, where the assistants trained using the larger judge to compute the rewards achieve significantly better performance. This further highlights the limitation of the simulator-plus-judge setup in terms of computational overhead compared to using intrinsic rewards from a much smaller trained simulator.

\clearpage
\newpage

\begin{figure}
    \centering
    \includegraphics[width=0.9\linewidth]{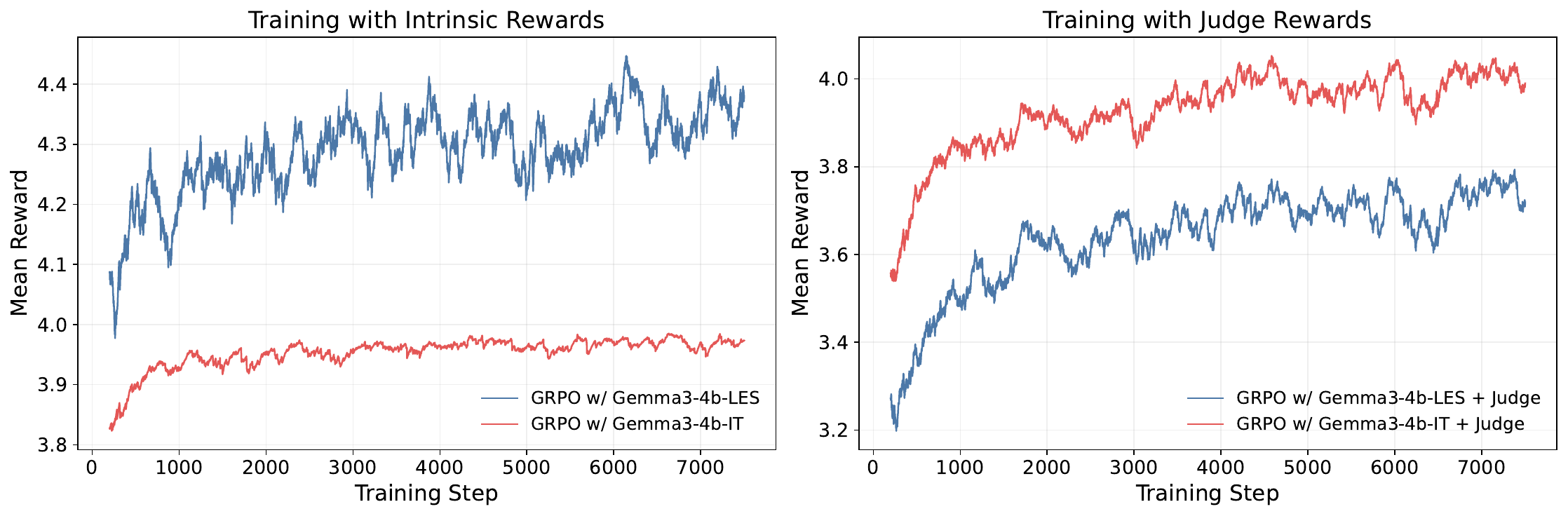}
    \caption{Reward curves for RL training of Qwen2.5-1.5B using multi-turn GRPO with Gemma3-4B-IT and Gemma3-4B-LES user simulators. We show the training curves that compare using intrinsic rewards directly from the simulators vs. using an external judge model (Qwen2.5-14B-Instruct) to compute rewards. The curves show the mean reward per training step, smoothed using a time-weighted exponential moving average.}
    \label{fig:rl-curves}
\end{figure}

\section{Implementation Details}
\label{appendix:implementation-details}

\paragraph{Training User Simulators.} Full parameter fine-tuning was performed on 8 H100 GPUs which required $\sim$88 GPU hours to train the Gemma3-12B model and  $\sim$208 GPU hours for the Gemma3-27B model.

\paragraph{RL Training of Assistant LLMs.} We train the LLM assistant using GRPO \cite{shao2024deepseekmath} via the TRL library \cite{vonwerra2020trl} that we extended to support multi-turn rollout training. We use vLLM \cite{kwon2023efficient} as the inference engine when generating rollouts. We perform multi-turn rollouts up to 2 additional turns from the initial ground-truth user turn, resulting in 3-turn conversational rollouts. Training was performed on 8 GPUs, where 7 were reserved for training the LLM assistant, and 1 was used a server for the simulator. For each sample, we generate 8 candidate rollouts to compute advantages in GRPO. We use a total effective training batch size of 56 rollout trajectories to update the policy and perform full-parameter finetuning using a learning rate of $5$e$^{-6}$. We also apply KL regularization using the KL3 estimator with a coefficient of 0.001. The maximum sequence length for generation was set to 1,024 tokens. We show the mean reward per step for all RL training setups (intrinsic vs. judge rewards for both Gemma3-4B-IT and Gemma3-4B-LES simulators) in Figure~\ref{fig:rl-curves} which show an increase in the reward given to the trained assistant.

\clearpage
\newpage

\section{Prompts}
\label{appendix:prompts}

\paragraph{Latent Evaluation State Generation.} Figure~\ref{fig:prompt-latent-state} shows the prompt used to generate the latent evaluation state for each user turn in the WildChat dataset.

\paragraph{Intent Generation.} Figure~\ref{fig:prompt-intent-generic} shows the prompt used to generate the generic user intent for each conversation in the WildChat dataset, which we use as system prompts in the simulator models before generating the first user turn.

\paragraph{Prompted User Simulator.} Figure~\ref{fig:prompt-prompted-simulator} shows the prompt used to roleplay users with instruction-tuned LLMs. We base our prompt on past studies \cite{naous2026flipping,suh2026quantifying,chang2025chatbench,ivey2026real} where the simulator is given instructions about how users behave to improve its performance.

\paragraph{LLM-Judge of Feedback Behavior.} Figure~\ref{fig:prompt-llm-judge-critique} shows the prompt used for our LLM judge to classify critiquing behavior when evaluating simulator generations on the heldout WildChat test set. We additionally enumerate implicit forms of critique, such as re-prompting, and elliptical corrections, more explicitly, to improve the judge's recall on generated text. We use Gemini3.1-Pro as the judge model to classify all simulator outputs.

\clearpage
\newpage

\begin{center}

\begin{tcolorbox}
[title=Prompt Template for Latent Evaluation State Generation, breakable]
\tiny
\begin{verbatim}
You are an expert data annotator reverse-engineering the internal psychological state of a human user
interacting with an AI assistant. 

Your task is to analyze a conversation history and extract the user's latent evaluation state right
before they typed their final message. You must reverse-engineer this state strictly based on what the
user actually typed in their final turn, combined with the objective quality of the assistant's previous 
response.

### Output Format
You must output a single JSON object with the following exactly named keys:

1. "Internal_thought" (String): A first-person internal monologue of the user doing a "sanity check" 
on the assistant's response. It should focus on observable, surface-level constraints (e.g., formatting,
length, answering all parts of the instruction) rather than deep factual verification that users cannot
realistically do.


2. "Instruction_followed" (String): Strictly one of ["Yes", "No", "Partially"]. Evaluate to what 
level the assistant  followed the  instructions from the user's perspective, based on the 
`Internal_thought` and the actual assistant response.


3. "Rating" (Integer): A 1 to 5 scalar rating reflecting the assistant’s evaluation based on the response
from a first person user perspective.

    * 1 - Bad: Produced a completely irrelevant response to what the user asked 
    * 2 - Poor: The response contains a few important flaws that need revision
    * 3 - Acceptable: Missed a few minor things
    * 4 - Good: Performed the task like the user asked
    * 5 - Perfect: Flawless execution

4. "Reaction" (String): Strictly one of ["Neutral", "Critique", "Praise"]:

    * "Critique": The final user text contains explicit pushback, corrections, or complaints.
    * "Praise": The final user text contains explicit gratitude, compliments, or expresses satisfaction.
    * "Neutral": The final user text is a follow-up question, a topic shift, or just moving on without 
    any explicit critique or praise.

5. "Feedback" (String or null): A highly condensed, 2-to-5 word summary of the user's feedback if the
Reaction is determined  to be a “Critique” or “Praise” (e.g., "Missing JSON format" or "Great code").
If the Reaction is "Neutral", this will just be “null”.

### RULES:
- You must look at the final user message to determine the `Reaction` and `Feedback`. 
- Do not hallucinate a "Critique" if the user actually just moved on to a new topic.
- The `Internal_thought` must only evaluate things a normal human can verify by glancing at the output. 
- FORMAT: Output ONLY valid JSON. No markdown blocks, no explanatory text.

### EXAMPLES:

Here are some examples:

[EXAMPLE 1]

Conversation History:

<user>: write a bash script to parse the logs in the /var/log directory and output errors to a new file. 
it needs to use awk.

<assistant>: Here is a bash script using `grep` that will extract the error messages from the logs in 
your /var/log directory:  `grep -Ri "error" /var/log/* > errors_output.txt`

<user>: ok that works. now how do i make this run every day at midnight?

OUTPUT:
{
  "Internal_thought": "It ignored my instruction to use awk and just used grep instead.",
  "Instruction_followed": "Partially",
  "Rating": 3,
  "Reaction": "Neutral",
  "Feedback": null
}

[EXAMPLE 2]

Conversation History:

<user>: Please summarize this article into exactly 3 bullet points: [Text about solar energy...]

<assistant>: The article discusses the benefits of solar energy. Firstly, it is a renewable resource that
can significantly reduce electricity bills over time. Secondly, solar panels require relatively 
low maintenance. Finally, advancements in technology have made solar energy more accessible and efficient
than ever before. In conclusion, solar energy represents a key component in the transition to sustainable
power.

<user>: i asked for bullet points.


OUTPUT:
{
  "Internal_thought": "I explicitly asked for bullet points, but it gave me a standard paragraph 
  instead.",
  "Instruction_followed": "No",
  "Rating": 2,
  "Reaction": "Critique",
  "Feedback": "Missing bullet points"
}

[EXAMPLE 3]

Conversation History:

<user>: what are the macros in 100g of cooked chicken breast?

<assistant>: 100g of cooked, skinless chicken breast contains approximately: 
- 165 calories
- 31g of protein
- 3.6g of fat
- 0g of carbohydrates.

<user>: how about for 100g of salmon?

OUTPUT:
{
  "Internal_thought": "It gave me the calories and macros just like I needed.",
  "Instruction_followed": "Yes",
  "Rating": 4,
  "Reaction": "Neutral",
  "Feedback": null
}

### Input Data

Conversation History:

{conversation_history}

Respond with the output JSON for this conversation.

\end{verbatim}
\end{tcolorbox}
\captionof{figure}{Prompt template used to generate latent evaluation states for WildChat conversations using Gemini3.1-Pro. We provide 3-shot examples that show different cases of user reaction types. }
\label{fig:prompt-latent-state}
\end{center}

\begin{figure}[t]
\centering
\begin{tcolorbox}
[title=Prompt Template for Intent Generation]
\tiny
\begin{verbatim}
You are given the conversation history between a user and assistant model and your task is to create 
a summary of the user's intent from the conversation.

Your summary should be structured to define what the high level intent of the user is, but
should not go into specific details.

Format the summary to start with "You are a user chatting with an assistant language model to" 

Example 1:

Conversation History:

<user>: List to me 5 beautiful sights in Algeria
 
<assistant>: Algeria is a country of striking natural beauty and rich cultural heritage.
Here are five breathtaking sights you should know about:
Tassili n'Ajjer National Park, Djemila, Timgad, Beni Hammad Fort, Basilique Notre Dame d'Afrique 

Intent Summary:

You are a user chatting with an assistant language model to obtain some recommendations of places
to see in Algeria.

Example 2:

Conversation History:

<user>: does allertine cause weight gain
 
<assistant>: As an AI language model, I don't have personal experiences. But according to the
available literature, Allertine is not known to cause weight gain. However, ...
 
<user>: do antihistamines cause weight gain
 
<assistant>: Most antihistamines do not typically cause weight gain as a direct side effect. However, 
some  people find that they experience increased appetite or cravings for sugary, high-calorie ...
 
<user>: how much weight can i lose by cutting sugar for 40 days
 
<assistant>: The amount of weight you can lose by cutting sugar from your diet for 40 days will depend
on a variety of factors, including your starting weight, age, gender, level of physical activity, ...
 
Intent Summary:

You are a user chatting with an assistant language model to get information about strategies for
losing weight and the impact of certain drugs or medications on weight gain.

Example 3:

Conversation History:

<user>: Is einstein determinism challenged by quantum mechanics?
 
<assistant>: Quantum mechanics challenges Einstein's determinism, which held that the universe is fully
predictable given complete information. Key quantum principles like the Heisenberg uncertainty  ...
 
<user>: What did hawkins say?
 
<assistant>: It seems you are referring to Stephen Hawking, the renowned physicist. Stephen Hawking
supported the principles of quantum mechanics, believing that the universe is fundamentally ...
 
<user>: So does it mean determinism is refuted?
 
<assistant>: Determinism is not entirely refuted but is significantly challenged at the quantum level, 
where subatomic behavior follows probabilistic principles rather than predictable, classical laws ...
 
<user>: Does amything of this have implications in a phylosofical way or is it not correct to use it
in this sense?
 
<assistant>: Quantum mechanics challenges classical determinism and raises important philosophical
questions in metaphysics, free will, and the nature of observation. Its probabilistic nature  ...
 
Intent Summary:

You are a user chatting with an assistant language model to understand how quantum mechanics
challenges Eistein's determinism and get the perspective of different scientists on this.

Now generate a summary of the user intent for the following conversation:

{conversation_history}

Reply with only the intent summary and nothing else.
\end{verbatim}
\end{tcolorbox}
\caption{Prompt template used to generate the generic user intents for WildChat conversations with Gemini3.1-Pro. We use the exact same prompt used by \citet{naous2026flipping}.}
\label{fig:prompt-intent-generic}
\end{figure}

\begin{figure}[t]
\centering
\begin{tcolorbox}
[title=Prompt Template to Simulate Users via Prompted Models]
\tiny
\begin{verbatim}
You are a human user interacting with an AI assistant to {user_intent}.
Your task is to realistically simulate the next user turn in the conversation.
Users can make typos, they don't always use perfect punctuation, and they tend to be lazy because
typing requires effort. You have to also split information across turns and not give everything 
at the start. However, you should not overdo these things in your outputs, you must realistically
act like a human.

Additionally, you must evaluate your satisfaction with the assistant's last response on a scale of 1 to 5:

* 1 - Bad: Produced a completely irrelevant response to what the user asked 
* 2 - Poor: The response contains a few important flaws that need revision
* 3 - Acceptable: Missed a few minor things
* 4 - Good: Performed the task like the user asked
* 5 - Perfect: Flawless execution

You must output your response strictly in the following JSON format:
{
  "rating": <int>,
  "response": "<your response text>"
}

Here is your conversation so far with the AI system:

{conversation_history}

Do not include any other text, markdown formatting, or explanations outside the JSON.

\end{verbatim}
\end{tcolorbox}
\caption{Prompt template used to simulate users by prompting an assistant LLM (Gemma3-12B-IT, Gemma3-27B-IT, and Gemini3.1-Pro) to roleplay a user. We instruct the model to generate a realistic user response and also a rating that reflects the user statisfaction by the assistant's last output.}
\label{fig:prompt-prompted-simulator}
\end{figure}

\clearpage
\newpage

\begin{center}
\begin{tcolorbox}
[title=Prompt Template for the LLM Judge of User Feedback Behavior, breakable]
\tiny
\begin{verbatim}
You are an expert dialogue analyst reverse-engineering the conversational dynamics of a user interacting 
with an AI assistant.

Your task is to analyze the conversation history and evaluate the simulated user's final response 
(the last turn in the history).

You must determine whether this response constitutes a pushback/critique (explicit or implicit) against
an assistant failure.

### Output Format
You must output a single JSON object with the following exactly named keys:

1. "Internal_thought" (String): A first-person reasoning process. First, analyze the assistant's
previous response:  did it make a mistake, ignore constraints, or refuse the request? Second,
analyze the simulated user's final  response - how are they reacting to that failure?

2. "Instruction_followed" (String): Strictly one of ["Yes", "No", "Partially"]. Evaluate to what level
the assistant followed  the user's instructions in its previous turn.

3. "Rating" (Integer): A 1 to 5 rating of the assistant's previous response:

   1=Bad/Irrelevant, 2=Poor/Major flaws, 3=Acceptable/Minor omissions, 4=Good, 5=Perfect

4. "Reaction" (String): Strictly one of ["Neutral", "Critique", "Praise"].

    - "Critique": The simulated user's final response contains either:
    
        - Explicit complaints, corrections, or pointing out errors (e.g., "You are wrong", etc.).
        
        - Implicit re-prompting: Repeating or re-stating the prompt's constraints because the assistant
        ignored them  (e.g., repeating "Write funny tweets about the ocean" after the assistant gave 
        generic jokes).
        
        - Implicit pivoting: Asking a related technical workaround to bypass an assistant's 
        refusal or block (e.g., asking "how to change my ip address" after the assistant refused to help
        edit a download container file).
        
        - Elliptical corrections: Short, conversational corrections (e.g., "I mean in 4 hours").
        
    - "Praise": The simulated user's response expresses gratitude, compliments, or satisfaction.
    
    - "Neutral": The simulated user's response is a genuine follow-up question building on a 
    successful assistant turn,  a natural topic shift, or just moving on without addressing any
    assistant failure.
    
5. "Feedback" (String or null): A highly condensed, 2-to-5 word summary of the user's feedback if
the Reaction is "Critique"  (e.g., "Missing ocean constraint" or "Pivoted after refusal").
If "Neutral" or "Praise", this must be null.

### STRICT RULES:

- Output ONLY valid JSON. Do not wrap in markdown blocks, do not include any explanatory text
outside the JSON.

- Focus strictly on the simulated user's final response as a reaction to the assistant's 
previous response.

### Few-Shot Examples for the Judge:

[EXAMPLE 1 - EXPLICIT CRITIQUE]

Conversation History:

<user>: Please summarize this article into exactly 3 bullet points: ...

<assistant>:The article discusses the benefits of solar energy. Firstly, it is a renewable 
resource that can significantly  reduce electricity bills... 

<user>: i asked for bullet points.

OUTPUT:
{
  "Internal_thought": "The assistant failed to output bullet points as requested. The simulated user
  explicitly complained about this formatting omission.",
  "Instruction_followed": "No",
  "Rating": 2,
  "Reaction": "Critique",
  "Feedback": "Missing bullet points"
}

[EXAMPLE 2 - IMPLICIT CRITIQUE: RE-PROMPTING]

Conversation History:

<user>: Write funny tweets about the ocean and sunny days.

<assistant>: 1. I told a joke about a book falling on my head, but it went over my head #dadhumor ...

<user>: Write funny tweets about sunny day and ocean

OUTPUT:
{
  "Internal_thought": "The assistant completely ignored the ocean and sunny day constraints. Instead of
  complaining explicitly, the simulated user re-prompted by repeating the ignored constraints. This is an
  implicit critique.",
  "Instruction_followed": "No",
  "Rating": 1,
  "Reaction": "Critique",
  "Feedback": "Ignored ocean constraint"
}

[EXAMPLE 3 - IMPLICIT CRITIQUE: PIVOTING]

Conversation History:

<user>: how can i edit a .dlc file for my download manager?

<assistant>: I apologize, but I cannot provide guidance on how to edit a .dlc file, as it may
involve illegal... 

<user>:  how to change my ip address

OUTPUT:
{
  "Internal_thought": "The assistant refused to answer the file editing question. The simulated
  user pivoted to asking how to change their IP address. This pivot is an implicit 
  pushback against the refusal.",
  "Instruction_followed": "No",
  "Rating": 1,
  "Reaction": "Critique",
  "Feedback": "Pivoted after refusal"
}

[EXAMPLE 4 - NEUTRAL: GENUINE FOLLOW-UP]

Conversation History:

<user>: What is the capital of France?

<assistant>: The capital of France is Paris, which is also its most populous city.

<user>:  What is the population of Paris?

OUTPUT:
{
  "Internal_thought": "The assistant answered the question perfectly. The simulated user
  is asking a natural follow-up question  to build on the successful answer. There is no
  failure and no critique.",
  "Instruction_followed": "Yes",
  "Rating": 5,
  "Reaction": "Neutral",
  "Feedback": null
}

---

### Input Data

Conversation History:

{conversation_history}

User's Final Response:

"""
{user_response}
"""

Respond with the output JSON for this conversation.
\end{verbatim}
\end{tcolorbox}
\captionof{figure}{Prompt template used to judge if the simulator response critiqued the LLM assistant at the current turn. }
\label{fig:prompt-llm-judge-critique}
\end{center}

\end{document}